# Performance Based Evaluation of Various Machine Learning Classification Techniques for Chronic Kidney Disease Diagnosis

Sahil Sharma
Department of Computer Science & IT
University Of Jammu
Jammu, India
E-mail: sahiil91@live.com

Vinod Sharma
Department of Computer Science & IT
University Of Jammu
Jammu, India
E-mail: vnodshrma@gmail.com

Atul Sharma
Department Of Internal Medicine
Government Medical College
Jammu, India
E-mail: dr.atul15@gmail.com

**Abstract:** Areas where Artificial Intelligence (AI) & related fields are finding their applications are increasing day by day, moving from core areas of computer science they are finding their applications in various other domains. In recent times Machine Learning i.e. a sub-domain of AI has been widely used in order to assist medical experts and doctors in the prediction, diagnosis and prognosis of various diseases and other medical disorders. In this manuscript the authors applied various machine learning algorithms to a problem in the domain of medical diagnosis and analyzed their efficiency in predicting the results. The problem selected for the study is the diagnosis of the Chronic Kidney Disease. The dataset used for the study consists of 400 instances and 24 attributes. The authors evaluated 12 classification techniques by applying them to the Chronic Kidney Disease data. In order to calculate efficiency, results of the prediction by candidate methods were compared with the actual medical results of the subject. The various metrics used for performance evaluation are predictive accuracy, precision, sensitivity and specificity. The results indicate that decision-tree performed best with nearly the accuracy of 98.6%, sensitivity of 0.9720, precision of 1 and specificity of 1.

## I. INTRODUCTION

Artificial Intelligence can be defined as *"the study and design of intelligent agents"* [1]. Programs which enable computers to function in the ways that make people seem intelligent are called artificial intelligent systems [2]. Machine learning which is a sub-domain of AI aims at providing various computational methods for imparting i.e. (accumulating, changing, and updating) knowledge to the intelligent systems. According to Ethem Alpaydin *"Machine learning is programming computers to optimize a performance criterion using example data or past experience. We have a model defined up to some parameters, and learning is the execution of a computer program to optimize the parameters of the model using the training data or past experience. The model may be predictive to make predictions in the future, or descriptive to gain knowledge from data, or both"* [3]. Machine learning can be seen as located at the intersection between computer science, applied mathematics and statistics sharing concepts with AI and information theory. The current desired contribution of AI in the field of medical science are the programs that can assist a medical expert in performing expert and more accurate diagnosis. These programs by making use of combination of various computational sciences find out patterns from the data used for training and then use these patterns in order to classify the test data into one of the possible categories.

Chronic Kidney Disease is a general term used for heterogeneous disorders affecting the structure and function of the kidney. It has a high mortality rate. The definition of Chronic Kidney Disease is based on the presence of kidney damage (i.e. albuminuria) or decreased kidney function for 3 months or more [4]. Kidney failure is traditionally regarded as the most serious outcome of Chronic Kidney Disease and symptoms are usually caused by complications of reduced kidney function. When symptoms are severe they can be treated only by dialysis and transplantation; kidney failure treated this way is known as end-stage renal disease. Other outcomes include complications of reduced GFR (glomerular filtration rate), such as increased risk of cardiovascular disease, acute kidney injury, infection, cognitive impairment, and impaired physical function [5]. Complications can occur at any stage, which often lead to death with no progression to kidney failure, & can arise from adverse effects of interventions to prevent or treat the disease [6].

The rest of the paper is organized as follows: In II[nd] section data used & various candidate techniques have been discussed.

III[rd] Section presents related work. Implementation of the methodology and Results have been analyzed and discussed in IV[th] section. Finally, V[th] section concludes the paper and describes future works.

## II. MATERIALS & METHODS

### A. Chronic Kidney Disease data

The dataset used in order to carry out the research in this manuscript has been obtained from the UCI data repository [7]. The dataset contains data of 400 people from the southern part of India with their ages ranging between 2-90 years. There are in total twenty four parameters, most of which are clinical in nature and the rest are physiological. Table 1 summarizes various parameters. In the preprocessing of the data the missing values were dealt with by replacing numeric and discrete integer values by attribute mean of the all the instances with the same class-label as that of the instance under consideration and nominal values were replaced using attribute mode.

**Table 1** Various parameters used and their allowed values.

| Parameter | Allowed Values |
|---|---|
| Age | Discrete Integer Values |
| Blood pressure | Discrete Integer Values |
| Specific gravity | Nominal Values (1.005,1.010,1.015,1.020,1.025) |
| Albumin | Nominal Values(0,1,2,3,4,5) |
| Sugar | Nominal Values(0,1,2,3,4,5) |
| Red blood cells | Nominal Values(Normal, Abnormal) |
| Pus cell | Nominal Values(Normal, Abnormal) |
| Pus cell clumps | Nominal Values(Present, Not-Present) |
| Bacteria | Nominal Values(Present, Not-Present) |
| Blood glucose random | Discrete Integer Values |
| Blood urea | Discrete Integer Values |
| Serum creatinine | Numeric Values |
| Sodium | Discrete Integer Values |
| Potassium | Numeric Values |
| Hemoglobin | Numeric Values |
| Packed cell volume | Discrete Integer Values |
| WBC count | Discrete Integer Values |
| RBC count | Numeric Values |
| Hypertension | Nominal Values(Yes, No) |
| Diabetes mellitus | Nominal Values(Yes, No) |
| Coronary artery disease | Nominal Values(Yes, No) |
| Appetite | Nominal Values(Good, Poor) |
| Pedal edema | Nominal Values(Yes, No) |
| Anemia | Nominal Values(Yes, No) |

### B. Candidate classification techniques

The authors analyzed twelve different classifiers majorly based on the following techniques: Decision-Tree, Support vector machine (SVM), Discriminant Analysis, K-nearest neighbor (KNN), Artificial neural networks (ANN). These techniques were selected for the analysis and study because of their popularity in the recent relevant literature. A brief description about the selected techniques has been given below:

*1) Decision Tree*: Decision tree classifiers classify data by making use of tree structure algorithms [8]. The underlying algorithm begins with the training samples and corresponding class labels. The training set is partitioned recursively based on a feature value into subsets. Each internal node represents a test on attribute; each edge (branch) represents an outcome of the test. A decision tree classifier identifies the class label of an unknown sample by following path root to the leaves, which represent the class label for that sample. The feature (attribute) i.e. selected as the root node is the one that best divides the training data. There are number of ways for finding the feature that best divides the training data, some of them are namely Information gain, myopic measures, G-statistics, chi-square, MDL etc. One cannot generalize any measure to be better than others. Any measure that results in a multiway tree (hence reduced complexity) and more balanced splits may be used depending on the dataset. Some of the commonly used decision tree algorithms are ID3, CART and C4.5.

*2) Support Vector Machine (SVM):* These classifiers are based on structural risk minimization principal and statistical learning theory with an aim of determining the hyperplanes (decision boundaries) that produce the efficient separation of classes. The underlying algorithm is Support Vector Classification (SVC) and it revolves around the perception of a "margin"- on either side of a hyperplane that divides two data classes. Maximizing the margin creates the largest possible distance among the hyperplane and the instances on either side of the hyperplane reduce an upper bound on the anticipated generalization error. It works on two types of data i.e. linearly separable data and linearly Non-separable data. In case of linearly separable data only one hyperplane is needed for separating the data but in the case of latter more than one hyperplanes are needed. Figure 2 depicts an example of a two-class problem with one separating hyperplane.

*3) Discriminant Analysis:* These classifiers work under the assumption that different classes generate data based on different Gaussian distributions. In the training phase the Gaussian distribution parameters for each class are estimated by the fitting function and in order to predict the classes (class-labels) of new data, the trained classifier finds the class with the smallest misclassification cost.

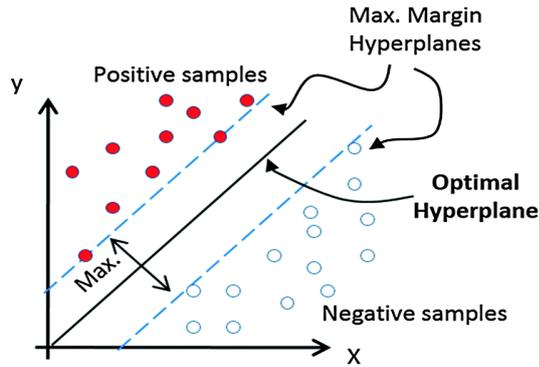

**Figure 1** An example of a two-class problem with one separating hyperplane.

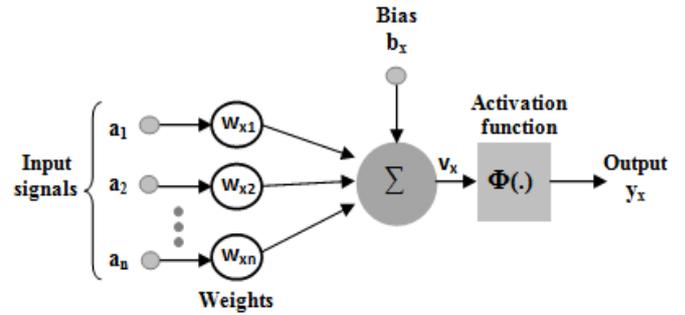

**Figure 2** An Artificial Neuron.

There are mainly two types of discriminant analysis classifiers namely – Linear Discriminant Analysis classifier (LDA) and Quadratic Discriminant Analysis classifier. The Quadratic Discriminant Analysis classifier can be considered as the generalization of LDA.

*4) K-Nearest Neighbors:* The k-nearest neighbors classifier is amongst the simplest of all machine learning algorithms. It is based on the principal that the samples that are similar lie in close proximity [9]. It classifies the test objects on the basis of number of closest training examples. It is also termed as a lazy- learning algorithm. KNN is a non parametric algorithm which means that it does not assume anything on the underlying data distribution. In this, the Euclidean distance is calculated between the test data and every sample in the training data followed by classifying the test data into a class in which most of k-closest neighbor's of training data belong to. K is usually a very small positive integer. As the value of K increases it becomes difficult to distinguish between the various classes. Cross-validation along with other heuristic techniques are used to choose an optimal value of K.

*5) Artificial Neural Network:* Artificial neural network (ANN) is a methodology inspired by the biological network of neurons. It is a powerful data-modeling tool capable of capturing, representing and simulating complex relationships between inputs and outputs by performing multiple parallel computations. These are analytical tools which try to emulate ''learning'' process of the cognitive system and the neurobiological functions of the human brain. In ANN, the neurons are grouped into different layers, an input layer, one or more hidden layers, and an output layer. Learning is achieved by repeatedly adjusting the numerical weights associated with the interconnecting edges between different artificial neurons. In addition to this an activation function is used that converts a neuron's weighted input to its output activation. Figure 2 illustrates a typical artificial neuron.

*C. Methodology*

For every individual classifier selected to train the system, first. Initially, all attributes corresponding to each case are described i.e. input data and out of these only one attribute is used to represent a decision for the given problem i.e. output data. Specific values are defined for the input attributes. The authors made use of 5-fold cross validation. Cross-Validation gives a good estimate of the accuracy of the final classifier trained with all the data. The procedure is:

1. Partition the data into k disjoint sets or folds
2. For each fold:
    a. Classifier is trained using the out-of-fold observations.
    b. Model performance is assessed using the in-fold data.
3. The average test error over all folds is calculated.

Afterward this methodology of each classifier differs but the last step i.e. common to all the classifiers is assigning a class to every single instance of the dataset.

## III. RELATED WORK

A number of researchers have used Machine Learning and data mining-based algorithms to solve problems in the field of medicine.

Abid Sarwar et al. [10] compared the accuracy of Naïve Bayes, artificial neural network, and KNN algorithm for the type II diabetes. Type II diabetes is a condition in which the pancreas is not able to produce the needed amount of insulin or the cell is not able to use the produced insulin (insulin resistance) which leads to abnormal glucose level in the blood. The results showed that Neural network with 96% prediction accuracy performs better than Naïve Bayes with 95% and KNN 91%.

Igor Kononenko [11] presented a view on the use of Machine learning techniques 1) for the interpretation of medical data

2) for intelligent analysis of medical data in the current scenario and 3) for assistance of physicians in diagnosis of

medical disorders, in the future. The authors suggested integration of machine learning techniques with the existing instrumentations for the acceptance of machine learning in medicine.

Yasodha et al. [12] did analysis of a database of diabetic patients using weka tool. The authors considered different algorithms such as REP Tree, Bayes Network, J48 and Random Tree classifiers for the study and compared the outputs. The main objective of the study was to develop a Diabetic expert system; inputs being patient's daily glucose rate and insulin dosages the system would predict the patient's insulin dosage for the next day.

T.Manju et al. [13] proposed a hybrid system of multi layer feed forward neural network (MLFFN) and genetic algorithm for assisting doctors in predicting the heart disease. The heart attack is the major cause of death in the world; its major causes are smoking, high blood pressure, unhealthy diet, obesity and diabetes. The ANN is trained using back propagation and feed forward neural network. Genetic algorithm is used for the weight optimization. The weights are associated with each connection in the neural network nodes. The data set consisted of 13 factors out of which only 6 were used for the purpose of training the ANN .The ANN then predicted the possibility of heart attack in a patient or not.

Babak Sokouti et.al. [14] proposed Levenberg–Marquardt (L-M) feedforward MLP neural network in order to classify cervical cell images obtained from 100 patients including healthy, low-grade intraepithelial squamous lesion and high-grade intraepithelial squamous lesion cases. The semi automated cervical cancer diagnosis system is composed of two phases: image pre processing/processing and feed forward MLP neural network. The results showed that cervical cell images were classified successfully with 100 % correct classification rate using the proposed method.

## IV. IMPLEMENTATION AND RESULTS

All the twelve classifiers were applied to the same dataset using MATLAB 2016a and the results were obtained and analyzed in the terms of predictive accuracy, sensitivity, precision & specificity. Predictive accuracy of Z% shows that the classifier is able to classify nearly Z% of instances correctly. Sensitivity can be defined as the ability to correctly detect cases which do have the condition. Mathematically it can be defined as the number of true positives divided by the number of true positives and false negatives. Specificity can be defined as the ability to correctly detect cases which do not have the condition. Mathematically, it can be expressed as the number of true negatives divided by the number of true negatives and false   positives. Precision is the number of correctly classified positive examples divided by the number of examples labeled by the system as positive [15] i.e. total number of true positives divided by the total number of true positives divided by the total number of true positives and false positives. True positives are the correctly recognized class examples, true negatives are correctly recognized examples that do not belong to the class, whereas false positives are examples that were incorrectly assigned to the class and false negatives are those examples that were not recognized as class examples [15].The maximum value of Sensitivity, Precision and Specificity can be 1 whereas their minimum value can be 0. The higher the value better is the performance of the classifier considered. Figure 3 shows the predictive accuracy of all the twelve classifiers whereas table 2 contains their sensitivity, precision and specificity values and their graphical representation is shown in figure 4.

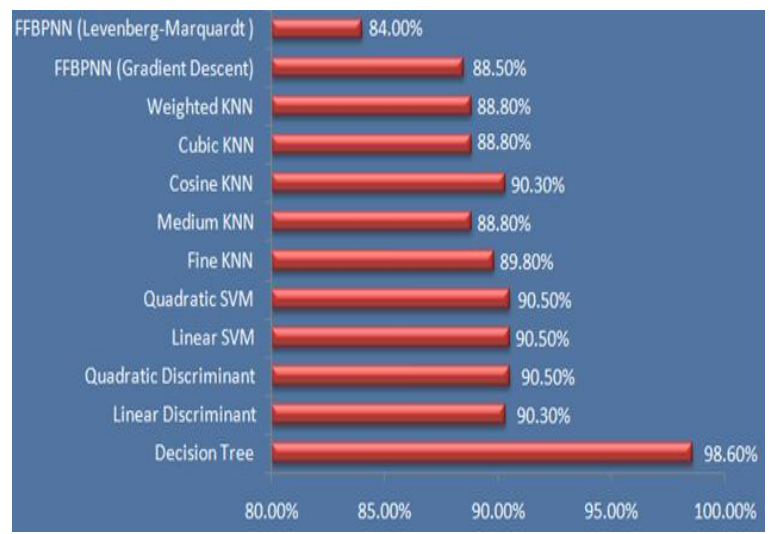

**Figure 3**   Predictive accuracy of all candidate classifiers.

Results show that decision trees performed best among all the candidate classifiers in terms of three out of four performance metrics with predictive accuracy of 98.60%, sensitivity of 0.9720, precision of 1 and specificity of 1. The three performance metrics in which it performed better than the rest of the candidate classifiers are predictive accuracy, specificity and precision. Linear Discriminant, Quadratic Discriminant, Linear SVM, Quadratic SVM, Fine KNN, Medium KNN, Cosine KNN, Cubic KNN, Weighted KNN, Feed Forward Back Propagation Neural Network using Gradient Descent (FFPBNN-Gradient Descent) performed better than the decision trees in terms of sensitivity. The overall performance of decision trees is better than all the other classifiers considered for this study.

**Table 2** Sensitivity, Precision and Specificity values of all candidate classifiers.

| Classifiers | Sensitivity | Precision | Specificity |
|---|---|---|---|
| Decision Tree | 0.9720 | 1 | 1 |
| Linear Discriminant | 0.9960 | 0.8736 | 0.7600 |
| Quadratic Discriminant | 1 | 0.8741 | 0.7600 |
| Linear SVM | 1 | 0.8741 | 0.7600 |
| Quadratic SVM | 1 | 0.8741 | 0.7600 |
| Fine KNN | 0.9920 | 0.8732 | 0.7600 |
| Medium KNN | 0.9760 | 0.8712 | 0.7600 |
| Cosine KNN | 0.9960 | 0.8736 | 0.7600 |
| Cubic KNN | 0.9760 | 0.8714 | 0.7600 |
| Weighted KNN | 0.9760 | 0.8714 | 0.7600 |
| FFBPNN (GD) | 0.9811 | 0.8320 | 0.7765 |
| FFBPNN (LM) | 0.9744 | 0.7640 | 0.7107 |

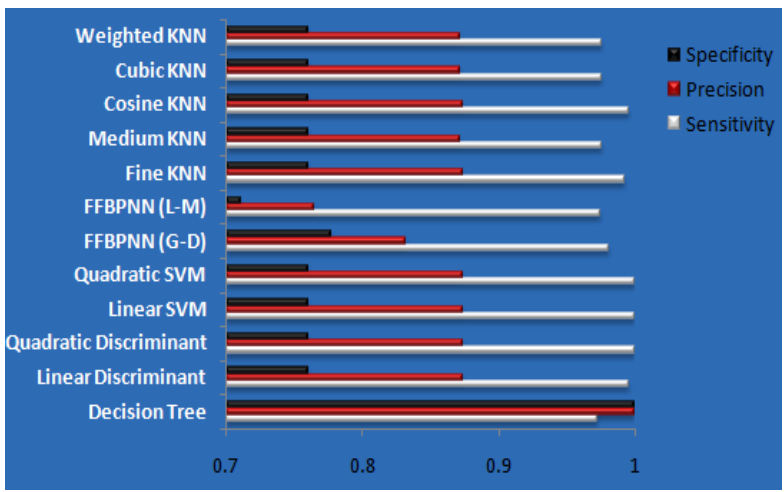

**Figure 4** Sensitivity, Precision and Specificity values of all candidate classifiers.

## V. CONCLUSION AND FUTURE SCOPE

Chronic Kidney disease is a generic term that covers various heterogeneous kidney disorders. Five to ten percent of the population worldwide suffers from this disease. Chronic Kidney Disease is a worldwide health crisis. A majority of the cases of Chronic Kidney Disease go undiagnosed or are diagnosed later in underdeveloped and developing nations; this is one of the prime reasons that higher percentage of these cases are from developing and underdeveloped nations as compared to developed nations where majority of people go through routine check-up and diagnosis. According to a report: "More than 80% of all patients who receive treatment for kidney failure are in affluent countries with universal access to health care & large elderly populations" [16].Machine learning based tools can be used for timely and accurate diagnosis of Chronic Kidney Disease by helping doctors in verifying the findings of their diagnosis in relatively short time thus helping a doctor to attend and diagnose more patients in less time as compared to the scenario where he/she has to go through the diagnosis process entirely manually. In the future course of this study one can try to further improve the two-class classification accuracy by evaluating some hybrid or ensemble techniques, in addition to this a subset of features can be extracted from the complete medical data-set of chronic kidney disease of twenty four parameters (features) without effecting the performance of the classification process, so that the financial burden a patient has to bear for undergoing various clinical tests can be reduced.

## AUTHOR'S BIOGRAPHY

**Sahil Sharma** holds a B.E in Computer Engineering and is currently pursuing his M.Tech in Computer Science (CS) from Department of Computer Science and IT, University of Jammu, J&K, India. His areas of interest include machine learning, data-mining & their applications.

**Dr. Vinod Sharma** is a professor of Computer Science and currently holds the position of Head, Department of Computer Science and IT, University Of Jammu, J&K, India. He Holds a PhD in Computer Science and has more than 25 publications in various journals and conferences. His research areas of interest are Artificial Intelligence (AI), Machine Learning, Databases and Data Mining.

**Dr. Atul Sharma** is pursuing his M.D in Internal Medicine & is currently a Resident Doctor at the GMC (Government Medical College), Jammu, J&K, India.